\documentclass[letterpaper]{article}  
\usepackage{aaai25}  
\usepackage{times}  
\usepackage{helvet}  
\usepackage{courier}  
\usepackage[hyphens]{url}  
\usepackage{graphicx} 
\usepackage{natbib}  
\usepackage{caption} 
\usepackage{algorithm}
\usepackage{algorithmicx}
\usepackage{epsfig}
\usepackage{color}
\usepackage{caption}
\usepackage{amsmath}
\usepackage{amsthm}
\usepackage{longtable}
\usepackage{rotating}
\usepackage{multirow}
\usepackage{url}
\usepackage{verbatim}
\usepackage{amssymb}
\usepackage{algpseudocode}
\usepackage{fancyhdr}    
\usepackage{subfigure}
\usepackage{booktabs}  
\usepackage{threeparttable} 
\usepackage{booktabs}
\usepackage{array}
\usepackage{ragged2e}
\usepackage{epstopdf}
\usepackage{arydshln}

\usepackage{newfloat}
\usepackage{listings}
\DeclareCaptionStyle{ruled}{labelfont=normalfont,labelsep=colon,strut=off} 
\floatstyle{ruled}
\newfloat{listing}{tb}{lst}{}
\floatname{listing}{Listing}
\title{FedMSGL: A Self-Expressive Hypergraph Based Federated Multi-View Learning}
\author {
    Daoyuan Li\textsuperscript{\rm 1},
    Zuyuan Yang\textsuperscript{\rm 1}\thanks{Corresponding author},
    Shengli Xie\textsuperscript{\rm 1}\textsuperscript{\rm 2}
}
\affiliations {
    \textsuperscript{\rm 1}School of Automation,\\ Guangdong Provincial Key Laboratory of Intelligent Systems and Optimization Integration,\\ Guangdong University of Technology\\
    \textsuperscript{\rm 2}Key Laboratory of iDetection and Manufacturing-IoT, Ministry of Education\\
    daoyuanli0914@aliyun.com, zuyuangdut@aliyun.com, shlxie@gdut.edu.cn
}

\usepackage{bibentry}

\begin{document}

\maketitle

\begin{abstract}
Federated learning is essential for enabling collaborative model training across decentralized data sources while preserving data privacy and security. This approach mitigates the risks associated with centralized data collection and addresses concerns related to data ownership and compliance. Despite significant advancements in federated learning algorithms that address communication bottlenecks and enhance privacy protection, existing works overlook the impact of differences in data feature dimensions, resulting in global models that disproportionately depend on participants with large feature dimensions. Additionally, current single-view federated learning methods fail to account for the unique characteristics of multi-view data, leading to suboptimal performance in processing such data. To address these issues, we propose a Self-expressive Hypergraph Based Federated Multi-view Learning method (FedMSGL). The proposed method leverages self-expressive character in the local training to learn uniform dimension subspace with latent sample relation. At the central side, an adaptive fusion technique is employed to generate the global model, while constructing a hypergraph from the learned global and view-specific subspace to capture intricate interconnections across views. Experiments on multi-view datasets with different feature dimensions validated the effectiveness of the proposed method.
\end{abstract}

%

\section{Introduction}

In the fast-paced world of modern technology, the proliferation of data is crucial for advancements in AI and machine learning. Traditional centralized data processing faces challenges due to the vast amount of distributed data and growing privacy and security concerns. Federated Learning (FL) offers a promising solution by leveraging the computational power and diverse data of edge devices\cite{Zhang2024FedTGPTG,Chen2024FedDATAA}. The framework of federated learning was introduced by Google in 2016\cite{google} and has since gained popularity in the industry.

Yang et al. systematically classified federated learning into three categories based on data partitioning\cite{TIST2019}: Horizontal Federated Learning (HFL), Vertical Federated Learning (VFL), and Federated Transfer Learning (FTL). HFL involves scenarios where data features are similarly distributed but the number of samples varies significantly\cite{HFL,HFL2,HFL3}. VFL occurs when the distribution of data features differs while the number of samples is similar\cite{VFL}. FTL applies to scenarios where data varies significantly in both feature space and sample count with limited overlap\cite{FTL1,FTL2}.

Vertical Federated Learning (VFL) has attracted researchers due to its practical applications\cite{VFL}. For instance, the FedBCD algorithm\cite{TSP2022} reduces communication costs by allowing local nodes to train independently, thus minimizing the need for frequent communication with the central server. The one-shot VFL method\cite{Sun2023} addresses communication bottlenecks and improves model performance with a limited number of overlapping samples. Additionally, the FedV framework\cite{FedV} employs functional encryption schemes to enhance privacy and security. Despite recent advances in VFL that have improved communication efficiency and privacy protection ability, these methods overlook two crucial issues: differences in feature dimensions and multi-view data.

Firstly, most existing VFL algorithms assume uniform feature space dimensions among participants or ignore disparities in feature dimensions\cite{TPDS2022}. In practical applications, data from large organizations tend to have more complex features, resulting in higher feature dimensions. These idealized assumptions lead to information bias, where the performance of the global model relies heavily on nodes with high-dimensional features, neglecting critical information from other nodes and deviating from the original intent of federated learning. Secondly, the pioneering work in VFL has primarily focused on single-view data, overlooking the consideration of multi-view data. With diverse data collection methods, data often consist of heterogeneous features capturing different views of samples. For example, in the Internet of Things (IoT) domain, multiple sensors are used for data collection\cite{IoT}, and in facial recognition, images are captured from different angles and under varying lighting conditions\cite{ORL}. Existing algorithms designed for single-view data struggle to capture the complex features of multi-view data\cite{FedMVL}. Multi-view learning methods aim to capture consistent and complementary information across different views. While centralized multi-view learning algorithms have been extensively studied\cite{MVLsurvey}, research on handling multi-view data in VFL environments is still in its early stages. Centralized multi-view approaches rely on accessing data from different views for consistent information. However, the requirement for privacy protection in a federated environment causes centralized methods to suffer from significant performance degradation or even become inapplicable. This issue has become a critical challenge that needs to be addressed in federated multi-view learning.

\begin{figure*}[ht]
    \centering
    \includegraphics[scale=0.54]{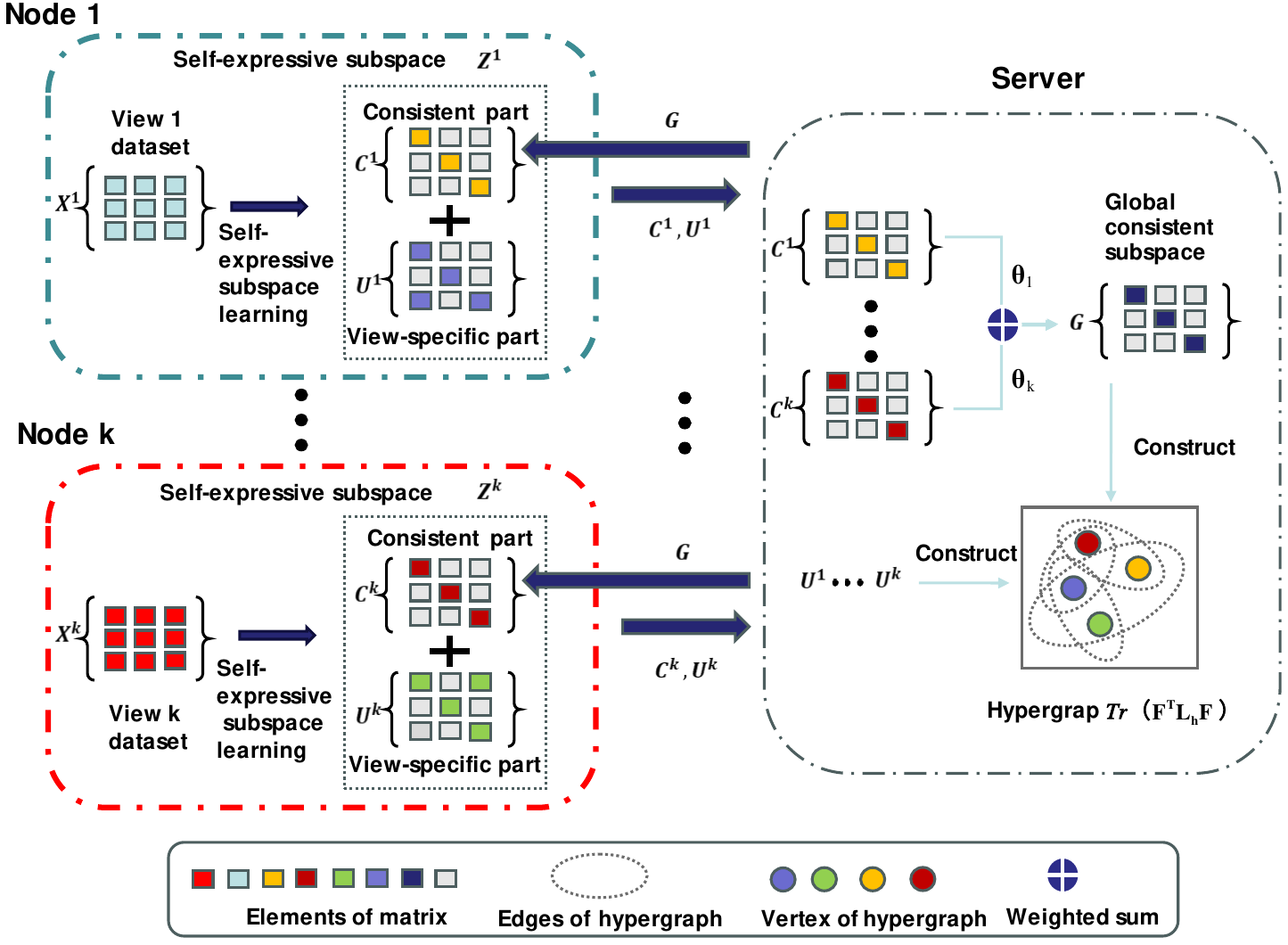}
    \caption{Framework of the proposed FedMSGL}
    \label{fig1}
\end{figure*}

Therefore, it is crucial to design an algorithm capable of feature dimension differences and multi-view data within the federated framework. We propose a novel vertical federated multi-view learning method called \textbf{Self-expressive Hypergraph Based Federated Multi-view Learning(FedMSGL)} to address the two challenges mentioned above. In the local node training process, we employ the self-expressive subspace learning technique to learn the latent representation with the same feature dimension and completed sample relation. Further, we divide the learned subspace into two parts to better explore the consistent and view-specific information of multi-view data within a federated environment. The adaptive integration strategy is applied at the central server to get the global consistent subspace. Then, we construct a hypergraph based on the global consistent and view-specific subspace to maximally capture potentially consistent information across views. The framework of the proposed FedMSGL is presented in Fig\ref{fig1}. The meanings of the notations used in this paper are given in Table\ref{notations}. The main innovations and contributions of our proposed method FedMSGL are summarized as follows.

\begin{itemize}
 
\item To address the impact of information bias caused by differences in data feature dimensions on federated learning algorithms, we employ a self-expressive subspace learning approach to obtain subspace embedding with the uniform feature dimension. This technique ensures that the learned global model treats all node feature information fairly.

\item We introduce a novel paradigm for federated learning algorithms in handling multi-view data. By constructing the optimal global hypergraph with consistent and view-specific subspace, we significantly enhance the clustering performance of federated multi-view learning algorithms. 

\item Our approach achieves performance on par with state-of-the-art centralized methods in multi-view datasets and demonstrates a notable improvement over existing federated multi-view methods.

\end{itemize}

\newcolumntype{L}[1]{>{\RaggedRight\arraybackslash}p{#1}}
\begin{table}[htp]
    \centering    
    \begin{tabular}{l c }
        \toprule
        Notations & Descriptions  \\
        \midrule
        $n$ & Number of samples \\
        $d$ & Feature dimension  \\
        $d_k$ & Feature dimension of $k$th view  \\
        $K$ & Number of views/nodes  \\
        $c$ & Number of cluster \\
        $\textbf{X}\in \mathbb{R}^{d \times n}$ & multi-view dataset  \\
        $\textbf{X}^k\in \mathbb{R}^{d_k \times n}$ & Dataset of $k$th view  \\
        $\textbf{C}^k\in \mathbb{R}^{n \times n}$ & Consistent part of $k$th view subspace   \\
        $\textbf{U}^k\in \mathbb{R}^{n \times n}$ & View-specific part of $k$th view subspace  \\
        $\textbf{M}^k\in \mathbb{R}^{n \times n}$ & Manifold coefficient matrix of $k$th view  \\
        $\textbf{G}\in \mathbb{R}^{n \times n}$ & Global consistent subspace  \\
        $\textbf{A}^k\in \mathbb{R}^{n \times n}$ & Affinity matrix of $k$th view \\
        $\textbf{A}\in \mathbb{R}^{n \times n}$ & Global affinity matrix\\
        $\textbf{F}\in \mathbb{R}^{n \times c}$ & Clustering indicator matrix\\
        $\|\cdot \|_F$ & Frobenius norm  \\
        $Tr()$ & Trace of the matrix\\
        \bottomrule
    \end{tabular}
    \caption{The Main Notations of FedMSGL}
    \label{notations}
\end{table}

\section{Related Work}
In this section, we provide a brief overview of relevant research advancements in federated learning and multi-view learning.
\subsection{Federated Learning}
As mentioned before, research efforts in federated learning can be categorized into three types: HFL, VFL and FTL. Considering the dynamic changes in node requirements within federated environments, Wang et al. propose an adaptive HFL framework to estimate the load variations at each node in power grid networks\cite{Wang2024}. Computationally efficient HFL has been a significant research focus. Liu et al. design a lightweight HFL algorithm based on top-k feature selection\cite{Liu2024}. Zhang et al. first combine the HFL method with virtual network embedding algorithm\cite{zhang2024}. He et al. observe that existing methods result in data wastage for the VFL field. By further leveraging dispersed features across nodes and employing data augmentation within nodes, they effectively extract valuable information from misaligned data\cite{he2024}. Zhu et al. extend the VFL framework to be applied in fuzzy clustering algorithms\cite{zhu2024}. Gao et al. employ complementary knowledge distillation techniques to enhance the robustness and security of VFL methods when facing stragglers and arbitrarily aligned data\cite{Gao_Wan_Fan_Yao_Yang_2024}. Yao et al. discover that existing methods struggle to defend against adversarial attacks and design attack techniques to identify vulnerabilities in current VFL algorithms\cite{yao2024constructing}. For the FTL, Wan et al. design a ring-based decentralized FTL framework, which enhances communication efficiency between clients through an adaptive communication mechanism\cite{WAN2024111288}. Qi et al. propose a differential privacy based knowledge transfer method within the federated environment\cite{qi2023differentially}.

\subsection{Multi-view Learning}
Multi-view learning methods obtain consensus representations by exploring cross-view consistency and complementarity in the data. Due to the strong interpretability of NMF (Non-negative Matrix Factorization) techniques, it is widely applied in multi-view learning. Huang et al. propose a one-step deep NMF method to produce a comprehensive multi-view representation\cite{DMF2024}. Cui et al. effectively improve the performance of NMF-based multi-view clustering methods by adding a fusion regularization term and utilizing partial label information\cite{cui2024}. Graph-based methods are another research hotspot. Tan et al. achieve a smoother multi-view consensus affinity graph representation by leveraging metric learning techniques\cite{Tan_Liu_Wu_Lv_Huang_2023}. Wang et al. innovatively leverage information across different dimensions to construct structural graphs and enhance the performance of multi-view graph learning\cite{sgl2024}. The subspace-based method is also an attractive branch. Chen et al. introduced anchor learning and globally guided local methods into multi-view subspace learning, enhancing the model's robustness to view discrepancy and computational efficiency\cite{FSMSC}. Additionally, Long et al. utilise tensors to further capture high-dimensional relationships, which facilitates learning a consensus subspace that preserves intricate geometric structures\cite{long2024}.

\section{Proposed Method}
Our research focuses on vertical federated scenarios. Consider a multi-view dataset $\textbf{X}=[\textbf{X}^1,\textbf{X}^2,\cdots,\textbf{X}^k ]\in \mathbb{R}^{d \times n}$ with $K$ views, each local node own its private sub-dataset $\textbf{X}^k\in \mathbb{R}^{d_k \times n}$ according to the vertical federated setting. Local nodes can only communicate with the central server, and raw data is unavailable for transmission to maintain privacy.

\subsection{Local Training}
Self-expression is a natural property of data, which refers to each data point within a union of subspace can be represented as an affine or linear combination of other data points in the same union\cite{TPAMI2013}. Assume that the private dataset in the $k$th node is $\textbf{X}^k \in \mathbb{R}^{d_k \times n}$, the self-expressive subspace learning method minimize the following function
\begin{equation}
\begin{split}
   &\quad \min_{\textbf{Z}^k} \|\textbf{X}^k-\textbf{X}^k\textbf{Z}^k \|_F^2 + \alpha \Gamma(\textbf{Z}^k),\\
   &\quad \text{s.t.}\quad\text{diag}(\textbf{Z}^k)=0,{(\textbf{Z}^{k})}^T\textbf{1}=\textbf{1}, \textbf{Z}^k \geq 0 
\end{split}
\end{equation}
where $\textbf{Z}^k$ is the subspace embedding of $\textbf{X}^k$, $\Gamma(\textbf{Z}^k)$ denotes the regularisation terms and $\alpha$ is a hyperparameter. The constraint $\text{diag}(\textbf{Z}^k)$ aims to avoid trivial solutions and ${(\textbf{Z}^k)}^T\textbf{1}=\textbf{1}$ indicates that data samples lie within a union of affine subspace. The subspace $\textbf{Z}^k$ learned by each node has the same dimension and effectively preserves latent sample relationships in the raw data.

To utilise consistent and complementary information from multi-view data within a federated learning framework, we partition the learned subspace into two components: a consistent part shared among views and a view-specific part. The objective function is given by
\begin{equation}
    \begin{split}
   &\quad \min_{\textbf{C}^k,\textbf{U}^k} \|\textbf{X}^k-\textbf{X}^k(\textbf{C}^k+\textbf{U}^k) \|_F^2 + \lambda_1\| \textbf{C}^k \|_F^2+ \lambda_3\| \textbf{U}^k \|_F^2,\\
   &\quad \text{s.t.} \quad \text{diag}(\textbf{C}^k)=0,\text{diag}(\textbf{U}^k)=0,(\textbf{C}^k+\textbf{U}^k)^T \textbf{1}=\textbf{1}, \\
   &\quad \quad \quad \textbf{C}^k \geq 0, \textbf{U}^k \geq 0
\end{split}
\end{equation}where $\textbf{C}^k$ denotes the consistent part of subspace, $\textbf{U}^k$ is view-specific part and $\lambda_1,\lambda_3>0$ are two trade-off hyperparameters. We introduce the Frobenius norm regularized term to encourage $\textbf{C}^k,\textbf{U}^k$ to preserve the group structure, which can bring high correlation samples closer together in their subspace embedding. Further, we add a coefficient matrix $\textbf{M}^k \in \mathbb{R}^{n \times n}$ to help $\textbf{C}^k$ to maintain the local manifold structure, facilitating the generation of hyperedges during central integration. For each data point, there exists a small neighbourhood where only the points from the same manifold lie approximately in a low-dimensional affine subspace\cite{manifold2011}. Based on the observation, a choice of the coefficient matrix $\textbf{M}^k = [\textbf{m}_1^k,\textbf{m}_2^k,\cdots,\textbf{m}_n^k]$ is 
\begin{equation}
  m_{ij}^k= 
   \begin{cases}
\frac{\|\textbf{x}^k_i-\textbf{x}^k_j \|_2}{\sum\limits_{t\neq i}\|\textbf{x}^k_t-\textbf{x}^k_j \|_2}, \text{if $i\neq j$. }   \\
0 \text{, if $i=j$.}
\end{cases}
\end{equation}The final objective of local training in $k$th node can be formulated as 
\begin{equation}
    \begin{split}
   &\quad \min_{\textbf{C}^k,\textbf{U}^k} \|\textbf{X}^k-\textbf{X}^k(\textbf{C}^k+\textbf{U}^k) \|_F^2 + \lambda_1\| \textbf{C}^k \|_F^2\\ &\quad \quad \quad +\lambda_2\|\textbf{M}^k\odot \textbf{C}^k \|_F^2 +\lambda_3\| \textbf{U}^k \|_F^2,\\
   &\quad \text{s.t.} \quad \text{diag}(\textbf{C}^k)=0,\text{diag}(\textbf{U}^k)=0,(\textbf{C}^k+\textbf{U}^k)^T \textbf{1}=\textbf{1}, \\
   &\quad \quad \quad \textbf{C}^k \geq 0, \textbf{U}^k \geq 0
\end{split}
\label{local object}
\end{equation}where $\odot$ represents element-wise multiplication and $\lambda_2$ is the hyperparameter.

\subsection{Central Integration}
We follow two intuitive assumptions to propose the adaptive integration strategy:(1) For multi-view data, the $k$th view subspace $\textbf{C}^k$ is a perturbation of the consistent subspace. (2)Subspace with high relevance to the global consistent subspace should be assigned large weights in the fusion process. The objective function of subspace fusion can be written as
\begin{equation}
    \min_{\textbf{G}}\sum\limits_{k=1}^K \theta_k \| \textbf{C}^k-\textbf{G} \|_F^2,
\label{fusion}    
\end{equation}
where $\textbf{G}$ denotes the global consistent subspace, $\textbf{C}^k$ is the consistent part of the uploaded subspace and $\theta_k$ is the adaptive weight. The weight of each view can be calculated with the natural index of inverse distance like
\begin{equation}
    \theta_k=\frac{1}{2 \text{exp}(\|\textbf{C}^k-\textbf{G} \|_F)}.
\label{theta}
\end{equation}

We further construct the global hypergraph with the learned consistent and view-specific subspace to capture complex relationships within different views. The affinity matrix of $k$th view is calculated by $\textbf{A}^k=\frac{(\textbf{G}+\textbf{G}^T)+(\textbf{U}^k+{(\textbf{U}^k)}^T)}{2}$ and the global affinity matrix is $\textbf{A}=\frac{1}{K}\sum\limits_{k=1}^K\textbf{A}^k$. Normally, we can construct the global $k$-NN similarity hypergraph from $\textbf{A}$. Let $\mathcal{H}=\{\mathcal{V},\mathcal{E} \}$ be a hypergraph,where $v_i\in\mathcal{V}$ is the set of vertices and $e_j\in\mathcal{E}$ is the set of hyperedges. Its incidence matrix $\textbf{H}\in \mathbf{R}^{|\mathcal{V}|\times|\mathcal{E}|}$ is 
\begin{equation}
    h_{(v_i,e_j)}=
    \begin{cases}
        1,\text{if}\quad v_i \in e_j \\
        0,\text{otherwise.}
    \end{cases}
\end{equation}
The normalized hypergraph Laplacian is defined as
\begin{equation}
    \textbf{L}_h = \textbf{I}-\textbf{D}_v^{(-1/2)}\textbf{H}\textbf{W}\textbf{D}_e^{(-1)}\textbf{H}^T\textbf{D}_v^{(-1/2)},
\end{equation}
where $\textbf{D}_v$ is vertex degree matrix, $\textbf{D}_e$ is hyperedge degree matrix and $\textbf{W}$ is the diagonal matrix of hyperedge weights(In this case, $\textbf{W}=\textbf{I}$). The elements in diagonal matrix $\textbf{D}_v$ is the degree of vertex $v_i$, which is the number of hyperedges it belongs to. And the elements in diagonal matrix $\textbf{D}_e$ is the degree of hyperedge $e_j$, which is the number of vertices in the hyperedge. Thus, the final objective function of the central integration stage can be defined as
\begin{equation}
    \begin{split}
    & \quad \min_{\textbf{G},\textbf{F}}\sum\limits_{k=1}^K \theta_k \| \textbf{C}^k-\textbf{G} \|_F^2+\beta\text{Tr}(\textbf{F}^T \textbf{L}_h\textbf{F}).\\
    & \quad \text{s.t.} \quad  \textbf{F}^T \textbf{F}= \textbf{I}
    \end{split}
    \label{centra}
\end{equation}
The $\textbf{F}\in \mathbb{R}^{n \times c} $ is the clustering indicator matrix, where $c$ is the cluster number. The $\beta$ is the trade-off parameter. Finally, the objective function of the proposed FedMSGL is summarized below:
\begin{equation}
    \begin{split}
   &\quad \min_{\textbf{C}^k,\textbf{U}^k,\textbf{G},\textbf{F}} \overbrace{\|\textbf{X}^k-\textbf{X}^k(\textbf{C}^k+\textbf{U}^k) \|_F^2 + \lambda_1\| \textbf{C}^k \|_F^2}^{\text{Local training}}+\\
   &\quad \quad \quad \quad \quad \overbrace{\lambda_2\|\textbf{M}^k\odot \textbf{C}^k \|_F^2+\lambda_3\| \textbf{U}^k \|_F^2}^{\text{Local training}}+\\
    &\quad\quad \quad \quad +\overbrace{\sum\limits_{k=1}^K \theta_k \| \textbf{C}^k-\textbf{G} \|_F^2+\beta\text{Tr}(\textbf{F}^T \textbf{L}_h\textbf{F})}^{\text{Central integration}}.\\
   &\quad  \text{s.t.} \quad \text{diag}(\textbf{C}^k)=0,\text{diag}(\textbf{U}^k)=0,(\textbf{C}^k+\textbf{U}^k)^T \textbf{1}=\textbf{1} \\
   &\quad \quad \quad \textbf{C}^k \geq 0, \textbf{U}^k \geq 0,\textbf{F}^T \textbf{F}= \textbf{I}
\end{split}
\label{FedMSGL}
\end{equation}

\subsection{Optimization}
The optimization goal of FedMSGL is to find the optimal value of Eq.\eqref{FedMSGL}. Unlike the optimization step in the centralized multi-view methods, the optimization in a federated environment is performed in two stages: first at the local nodes, and then at the central server. 

\subsubsection{Local nodes}
The optimization object of the local $k$th node is expressed as Eq.\ref{local object}. The optimal solution can be obtained by solving two subproblems.

\textit{1.$\textbf{C}^k$ Updating} 

When fixing the $\textbf{U}^k$, the sub-optimisation object of Eq.\eqref{local object} is written as
\begin{equation}
    \begin{split}
   &\quad \min_{\textbf{C}^k} \|\textbf{X}^k-\textbf{X}^k(\textbf{C}^k+\textbf{U}^k) \|_F^2 + \lambda_1\| \textbf{C}^k \|_F^2\\ &\quad \quad \quad +\lambda_2\|\textbf{M}^k\odot \textbf{C}^k \|_F^2.\\
   &\quad \text{s.t.} \quad \text{diag}(\textbf{C}^k)=0,(\textbf{C}^k+\textbf{U}^k)^T \textbf{1}=\textbf{1}, \\
   &\quad \quad \quad \textbf{C}^k \geq 0
\end{split}
\label{l1}
\end{equation}
We denote the latent solution of $\textbf{C}^k$ without the constraints as $\Tilde{\textbf{C}}^k$. The unconstrained objective function of Eq.\eqref{l1} can be formed as

\begin{equation}
    \begin{split}
   &\quad \min_{\Tilde{\textbf{C}}^k}\|\textbf{X}^k-\textbf{X}^k(\Tilde{\textbf{C}}^k+\textbf{U}^k) \|_F^2 + \lambda_1\| \Tilde{\textbf{C}}^k \|_F^2\\ 
   &\quad \quad \quad +\lambda_2\|\textbf{M}^k\odot \Tilde{\textbf{C}}^k \|_F^2.
\end{split}
\end{equation}
Let the partial derivative w.r.t $ \Tilde{\textbf{C}}^k$ equal to 0 ,the closed-form solution of $\Tilde{\textbf{C}}^k$ is
\begin{equation}
\begin{split}
     &\Tilde{\textbf{C}}^k=(\lambda_1 \textbf{I}+\lambda_2\text{diag}(\textbf{M}^{k^2})+{(\textbf{X}^k)}^T\textbf{X}^k)^{-1}{(\textbf{X}^k)}^T\\
     &(\textbf{X}^k-\textbf{X}^k\textbf{U}^k).
\end{split}
\end{equation}
Then the optimal problem in Eq.\eqref{l1} can be rewritten as
\begin{equation}
\begin{split}
   & \min_{c^k_{ij}}\sum\limits_{i=1}^n\sum\limits_{j=1}^n(c^k_{ij}-\Tilde{c}^k_{ij} )^2.\\
    &  \text{s.t.} \quad c^k_{ii}=0,(\textbf{c}^k_{i}+\textbf{u}^k_{i})^T \textbf{1}=1,  \textbf{c}^k_{i} \geq 0
\end{split}
\label{ct2}
\end{equation}
Considering that Eq.\eqref{ct2} is independent for each row, thus we can update each row of $\textbf{C}^k$ by solving the following equation
\begin{equation}
\begin{split}
   &  \quad \min_{\textbf{c}^k_i}\|\textbf{c}^k_i-\Tilde{\textbf{c}}^k_i \|_2^2.\\
    &\text{s.t.} \quad c^k_{ii}=0,(\textbf{c}^k_{i}+\textbf{u}^k_{i})^T \textbf{1}=1, \textbf{c}^k_{i} \geq 0
\end{split} 
\end{equation}
Its augmented Lagrangian function is formulated as
\begin{equation}
\begin{split}
&\mathcal{L}^k(\textbf{c}_i^k,\phi_1^k,\phi_2^k)=\|\textbf{c}_i^k-\Tilde{\textbf{c}_i}^k \|_F^2-\phi_1^k((\textbf{c}^k+\textbf{u}^k)^T \textbf{1}-1)\\
&-{({\phi_2^k})}^T\textbf{c}^k_i,
\end{split}
\end{equation}
where $\phi_1^k,\phi_2^k$ are Lagrangian multipliers. Using Karush-Kuhn-Tucker conditions and letting the $\frac{\partial\mathcal{L}^k}{\partial \textbf{c}^k_i}=0$, we have
\begin{equation}
\begin{split}
    &\textbf{c}^k_i = \max(\Tilde{\textbf{c}}^k_i+\phi_1^k\textbf{1},0),\\
    & \phi_{1,ij}^k =\sum\limits_{i=1}^n\frac{1-\sum\limits_{j=1}^n(\Tilde{c}^k_{ij}+u^k_{ij})}{n}. 
\end{split}\label{cupdate}
\end{equation}

\textit{2.$\textbf{U}^k$  Updating} 

When fixing the $\textbf{C}^k$ and other irrelevant terms, the sub-optimisation problem of Eq.\eqref{local object} is transformed to
\begin{equation}
    \begin{split}
   &\quad \min_{\textbf{U}^k}\|\textbf{X}^k-\textbf{X}^k(\textbf{C}^k+\textbf{U}^k) \|_F^2 + \lambda_3\| \textbf{U}^k \|_F^2.\\ 
   &\quad \text{s.t.} \quad \text{diag}(\textbf{U}^k)=0,(\textbf{C}^k+\textbf{U}^k)^T \textbf{1}=\textbf{1}, \\
   &\quad \quad \quad \textbf{U}^k \geq 0
\end{split}
\label{l2}
\end{equation}
The updating process of $\textbf{U}^k$ is similar with $\textbf{C}^k$, we directly give the optimal solution and the details are presented in the appendix. The optimal $\Tilde{\textbf{U}^k}$ is calculated by
\begin{equation}
    \Tilde{\textbf{U}}^k=(\lambda_3\textbf{I}+{(\textbf{X}^k)}^T\textbf{X}^k)^{-1}{(\textbf{X}^k)}^T(\textbf{X}^k-\textbf{X}^k\textbf{C}^k).
\end{equation}
And the optimal $\textbf{u}^k_i$ blows to
\begin{equation}
\begin{split}
    &\textbf{u}^k_i = \max(\Tilde{\textbf{u}}^k_i+\phi_1^k\textbf{1},0). 
\end{split}\label{uupdate}
\end{equation}
After solving the two sub-problems above, the $\textbf{C}^k$ and $\textbf{U}^k$ are transferred to the central server for integration.

\begin{algorithm}[ht]
\caption{FedMSGL}
\label{alg1}
\textbf{Input}: Multi-view dataset with $n$ view ${\bf X}^{1}$, ${\bf X}^{2}$, $\cdots$, ${\bf X}^{n}$, Hyperparameter $\lambda_1$,$\lambda_2$ and $\lambda_3$.\\
\textbf{Ouput}: Global model $\textbf{G}$ and the cluster indicator matrix $\textbf{F}$.\\
\textbf{Initialize} $\textbf{C}^k $, $\textbf{U}^k$ \begin{algorithmic}[1]
\While{converge}
\Statex\textbf{Local node:} do calculation among nodes parallel
\For{ $i = 1$  \textbf{to}  $itermax $ }
\If{$i \neq 1$}
    \Statex\quad\quad\quad replace $\textbf{C}^k$ with $\textbf{G}$;
\Else
    \Statex\quad\quad\quad update $\textbf{C}^k$ by Eq.\eqref{cupdate};\\ 
    \quad\quad\quad update $\textbf{U}^k$ by Eq.\eqref{uupdate}; 
\EndIf
\EndFor
\Statex  \quad\quad send $\textbf{C}^k$ and $\textbf{U}^k$ to server;
\Statex\textbf{Central Server:} 
\For{ $i = 1$  \textbf{to}  $ itermax $ }
    \Statex\quad\quad\quad update $\textbf{G}$ by Eq.\eqref{Gupdate};\\
 \quad\quad\quad   update $\textbf{F}$ by solving Eq.\eqref{Fupdate};
\EndFor
    \Statex \quad\quad send $\textbf{G}$ to nodes;
\EndWhile
\end{algorithmic}
\end{algorithm}

\subsubsection{Central server} The optimization process is equal to solving two sub-problems.

\textit{3.$\textbf{G}$ Updating}

Noting that $\textbf{L}_h$ is a function of $\textbf{G}$. Thus, when fixing $\textbf{F}$ and $\textbf{C}^k$,the objective function can be rewritten as  
\begin{equation}
\min_{\textbf{G}}\sum\limits_ {k=1}^n\theta_k\| \textbf{C}^k - \textbf{G}  \|_F^2 +\beta Tr(\textbf{F}^T\textbf{L}_h\textbf{F}),
\label{l3}
\end{equation}
where $\theta_k$ is the adaptive parameter which is calculated by Eq.\eqref{theta}. Additionally, the second term in Eq.\eqref{l3} is equal to
\begin{equation}
    Tr(\textbf{F}^T\textbf{L}_h\textbf{F})=\sum\limits_{i=1}^n\sum\limits_{j=1}^n\frac{1}{2}\|\textbf{f}_i-\textbf{f}_j \|^2g_{ij},
\end{equation}where $\textbf{f}$ denotes the row vector of $\textbf{F}$. 
To solve the sub-promblem in Eq.\eqref{l3}, we introudce $\textbf{z}_i \in \mathbb{R}^{n\times1}$ as an auxiliary variable. The $j$th entry of $\textbf{z}_i$ is $ z_{ij} = \| \textbf{f}_i-\textbf{f}_j  \|^2 $. Then, we solve the Eq.\eqref{l3} column-wisely
\begin{equation}
     \min_{\textbf{G}(:,i)}\sum\limits_{k=1}^n\theta_k\|\textbf{C}^k(:,i)-\textbf{G}(:,i) \|^2+ \frac{\beta}{2}\textbf{z}_i^T\textbf{G}(:,i),
\end{equation}
and letting its derivative w.r.t $\textbf{G}(:,i)$ equal to zero, it yields 
\begin{equation}
    \textbf{G}(:,i)=\frac{\sum_{k=1}^n\theta_k\textbf{C}^k(:,i)-\frac{\beta\textbf{z}_i}{4}}{\sum_{k=1}^n\theta_k}.
    \label{Gupdate}
\end{equation}

\textit{4.$\textbf{F}$ Updating}

When fixing the $\textbf{G}$, the objective function is written as
\begin{equation}
    \begin{split}
         &\quad\min_{\textbf{F}}Tr(\textbf{F}^T\textbf{L}_h\textbf{F}).\\
         &\quad \text{s.t.}\quad \textbf{F}^T\textbf{F} = \textbf{I}
    \label{Fupdate}
    \end{split}
\end{equation}The optimal solution of indicator matrix $\textbf{F}$ is derived from $c$ eigenvectors of $\textbf{L}_h$ corresponding to the $c$ smallest eigenvalues. The detailed explanation is presented in the appendix. 

When the integration process finishes, the learned global model $\textbf{G}$ is transferred to each node and replaces $\textbf{C}^k$ before a new round of local training. The pseudocode of FedMSGL is summarized in Algorithm 1. The complex and convergence analyses are presented in the appendix.

\begin{figure}[htb]
\centering
\subfigure[$\lambda_2$]{
		\includegraphics[scale=0.25]{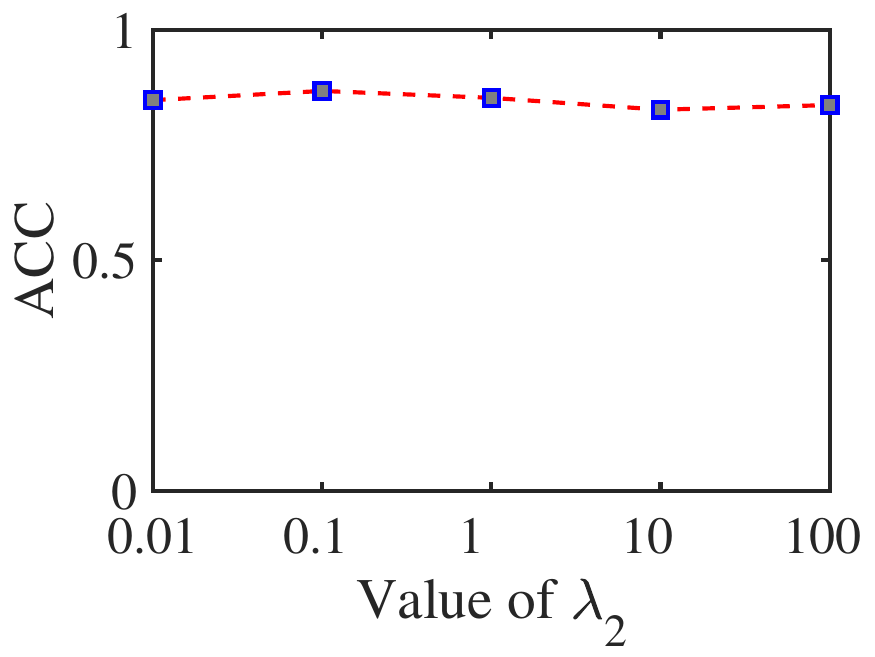}}
\subfigure[$\lambda_1 $ and $ \lambda_3$]{
		\includegraphics[scale=0.22]{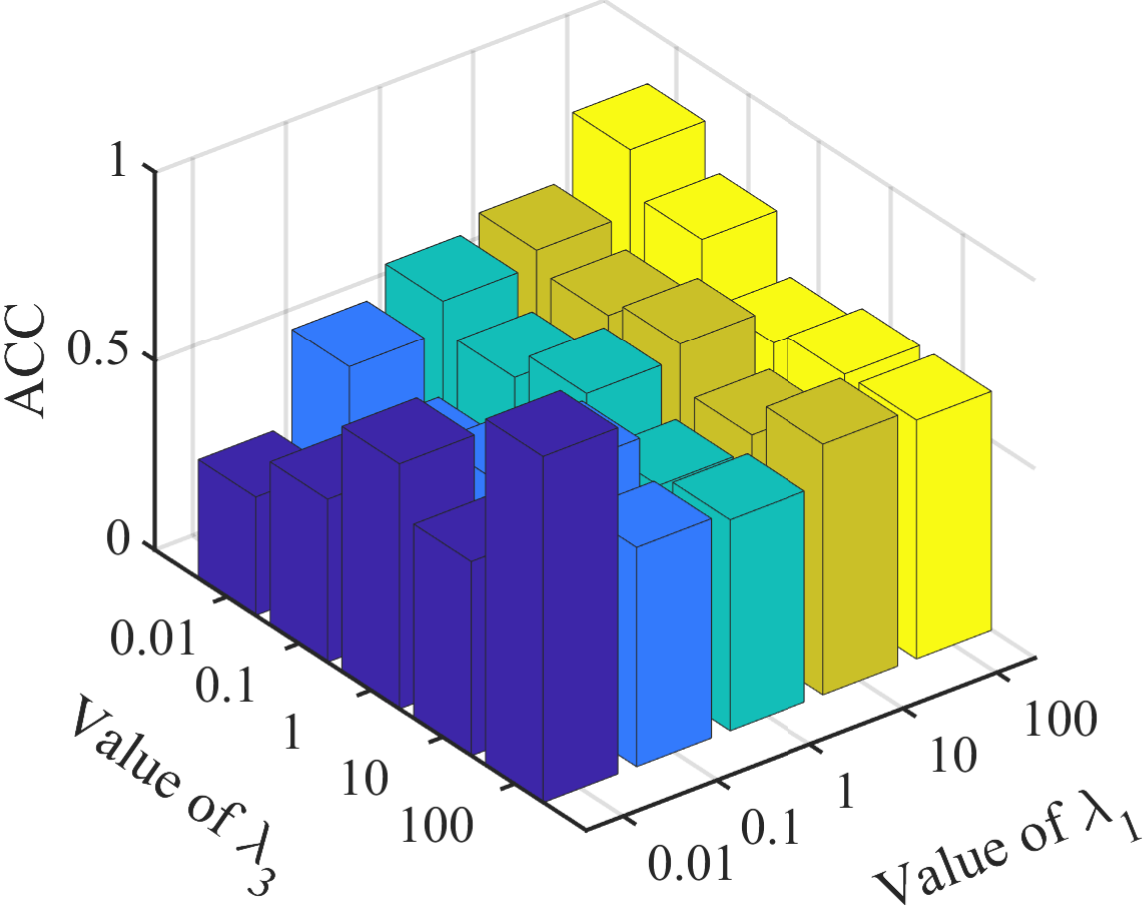}}
\subfigure[$\beta$]{
		\includegraphics[scale=0.25]{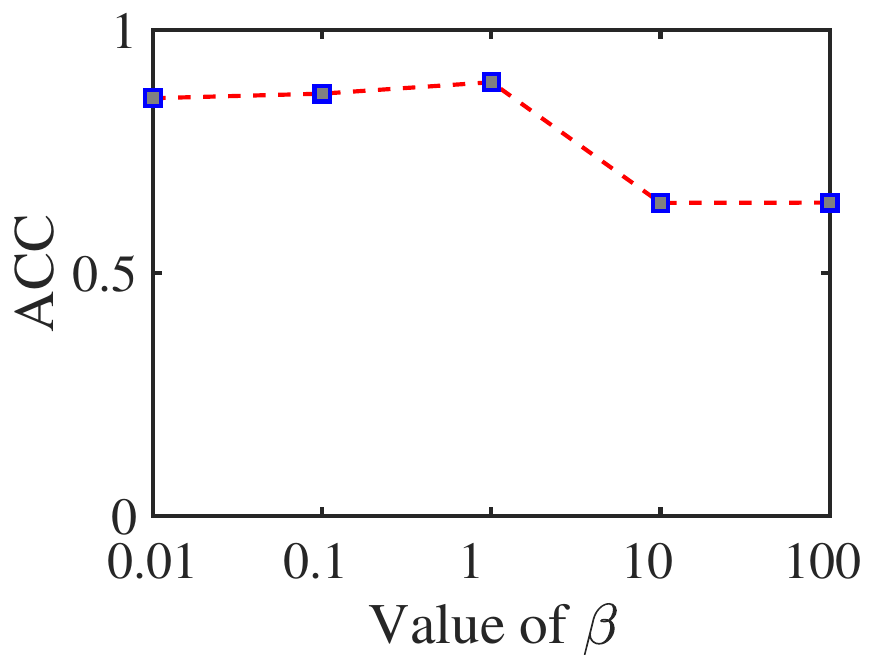}}
\subfigure[Convergence curve on BBC Sport]{
		\includegraphics[scale=0.30]{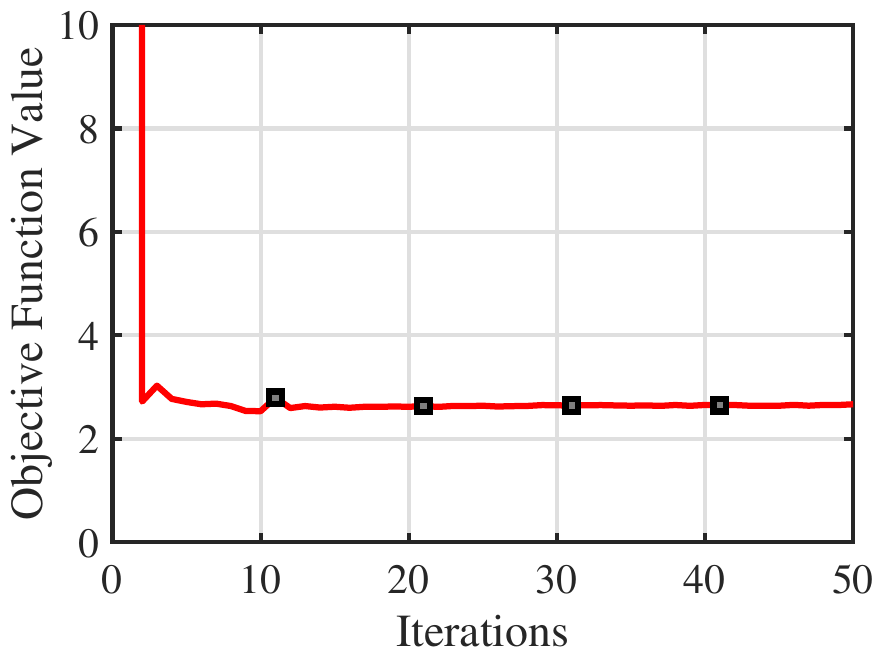}}
\caption{Performances change and convergence curve on BBC Sport dataset. Performances change on other datasets are given in the appendix.}
\label{ps}
\end{figure}

\section{Experiments}
\subsection{Datasets and Compared Methods}
We conducted experiments on five multi-view datasets and four have differing feature dimensions. {\bf Sonar\footnote[1]{http://archive.ics.uci.edu/dataset/151/connectionist+bench
+sonar+mines+vs+rocks}},{\bf BBC Sport\footnote[2]{http://mlg.ucd.ie/datasets/bbc.html}},{\bf ORL\footnote[3]{https://www.cl.cam.ac.uk/research/dtg/attarchive/facedat
abase.html}} ,{\bf Handwritten\footnote[4]{https://archive.ics.uci.edu/dataset/72/multiple+features}} and {\bf Caltech101-7\footnote[5]{http://www.vision.caltech.edu/archive.html}}. The statistics of the five datasets are summarized in Table.\ref{dataset}. We compare the proposed method with five centralised multi-view learning methods including \textbf{SC}(Spectral clustering, baseline),\textbf{SFMC}\cite{SFMC},\textbf{CDMGC}\cite{CDMGC}, \textbf{FSMSC}\cite{FSMSC} and \textbf{RCAGL}\cite{RCAGL2024}. And three federated multi-view learning methods like \textbf{FedMVL(2022)}\cite{FedMVL}, \textbf{FedMVFCM}\cite{Hu2023} and \textbf{FedMVFPC}\cite{Hu2023}. Details of the datasets and compared methods are presented in the appendix.

\setlength{\tabcolsep}{4pt}
\begin{table}[H]
\centering
\begin{tabular}{cccc}
\toprule
Dataset & sample &view &feature dimension \\
\midrule
Sonar&208&2&60/60 \\
BBC Sport&544&2&3283/3183 \\
ORL&400&3&4096/3304/6750 \\
Handwritten&2000&6&76/216/64/240/47/6 \\
Caltech101-7&1474&7&254/48/512/1984/928/40 \\
\bottomrule
\end{tabular}
\caption{Statistics of five real-world datasets}
\label{dataset}
\end{table}

\begin{table*}[htb]
\centering
\setlength{\tabcolsep}{1mm}
\begin{tabular}{ccccccc}
\toprule
Methods  &Sonar &BBC Sport &ORL &Handwritten &Caltech101-7 \\
\midrule
&&&ACC\\
\midrule%
SC(mean)& 0.5673$\pm$0.00 &0.5423$\pm$0.00 &0.6600$\pm$0.00 &0.2161$\pm$0.00 &0.5909$\pm$0.00 \\
SFMC& 0.5144$\pm$0.00 &0.3658$\pm$0.00 &0.7325$\pm$0.00 &0.8565$\pm$0.00 &0.6493$\pm$0.00 \\
CDMGC & 0.5427$\pm$0.00 &0.7353$\pm$0.00 &0.7900$\pm$0.01 &0.8445$\pm$0.01 &\underline{0.7631}$\pm$0.05 \\
FSMSC & 0.5721$\pm$0.00 &\underline{0.8272}$\pm$0.00 &\underline{0.8025}$\pm$0.00 &\underline{0.8845}$\pm$0.00 &0.6425$\pm$0.00\\
RCAGL & 0.5625$\pm$0.00 &0.6011$\pm$0.00 &0.7100$\pm$0.00 &0.8775$\pm$0.00 &\underline{0.7327}$\pm$0.00\\
\hdashline
FedMVL& \underline{0.6250}$\pm$0.00 &0.5662$\pm$0.03 &0.5600$\pm$0.03 &0.5245$\pm$0.03 &0.4770$\pm$0.05\\
FedMVFCM& 0.5154$\pm$0.00 &0.5257 $\pm$0.00 &0.5715 $\pm$0.00 &0.7115 $\pm$0.00 &0.6705 $\pm$0.00\\
FedMVFPC& 0.5279$\pm$0.01 &0.5468$\pm$0.09 &0.6525$\pm$0.02 &0.7365$\pm$0.05 &0.7162$\pm$0.05\\
FedMSGL& \underline{0.6466}$\pm$0.00 &\underline{0.9022}$\pm$0.04 &\underline{0.8283}$\pm$0.02 &\underline{0.8898}$\pm$0.08 &0.7256$\pm$0.07 \\
\bottomrule
\toprule
&&&Purity\\
\midrule
SC(mean)& 0.5721$\pm$0.00 &0.5625$\pm$0.00 &0.6958$\pm$0.00 &0.2241$\pm$0.00 &0.6122$\pm$0.00 \\
SFMC& \underline{0.6826}$\pm$0.00 &0.6047$\pm$0.00 &0.8275$\pm$0.00 & 0.8775$\pm$0.00 &0.7856$\pm$0.00 \\
CDMGC & 0.5421$\pm$0.00 &0.7592$\pm$0.00 &\underline{0.8600}$\pm$0.01 &0.8820$\pm$0.00 &\underline{0.8937}$\pm$0.09 \\
FSMSC & 0.5923$\pm$0.00 &\underline{0.8327} $\pm$0.00 &0.8275$\pm$0.00 &0.8845$\pm$0.00 &\underline{0.8731}$\pm$0.00\\
RCAGL & 0.6336$\pm$0.00 &0.8217$\pm$0.00 &0.8275$\pm$0.00 &0.8775$\pm$0.00 &0.7856$\pm$0.00\\
\hdashline
FedMVL& 0.6250$\pm$0.00 &0.5993$\pm$0.03 &0.6025$\pm$0.01 &0.5190$\pm$0.00 &0.5417$\pm$0.00\\
FedMVFCM& 0.6154$\pm$0.00 &0.5276 $\pm$0.00 &0.6325$\pm$0.00 &0.7216 $\pm$0.00 &0.5522 $\pm$0.00\\
FedMVFPC& 0.5375$\pm$0.00 &0.6222$\pm$0.09 &0.6971$\pm$0.01 &0.7965$\pm$0.04 &0.7353$\pm$0.05\\
FedMSGL& \underline{0.6466}$\pm$0.00 &\underline{0.9061}$\pm$0.02 &\underline{0.8520}$\pm$0.01 &\underline{0.8964}$\pm$0.08 &0.8453$\pm$0.01 \\
\bottomrule
\toprule
&&&NMI\\
\midrule
SC(mean)& 0.0165$\pm$0.00 &0.2438$\pm$0.00 &0.7709$\pm$0.00 &0.1290$\pm$0.00 &0.0864$\pm$0.00 \\
SFMC& 0.0312$\pm$0.00 &0.0334$\pm$0.00 &\underline{0.8922}$\pm$0.00 &\underline{0.9047}$\pm$0.00 & 0.5096$\pm$0.00 \\
CDMGC & 0.0642$\pm$0.00 &0.6933$\pm$0.00 &0.8411$\pm$0.02 &0.8867$\pm$0.00 &\underline{0.6247}$\pm$0.12 \\
FSMSC & 0.0146$\pm$0.00 &\underline{0.7217}$\pm$0.00 &0.8909$\pm$0.00 &0.8011$\pm$0.00 &0.5141$\pm$0.00\\
RCAGL & 0.0256$\pm$0.00 &0.4515$\pm$0.00 &0.8850$\pm$0.00 &0.8061$\pm$0.00 &\underline{0.6394}$\pm$0.00\\
\hdashline
FedMVL& \underline{0.1398}$\pm$0.00 &0.3832$\pm$0.03 &0.7408$\pm$0.01 &0.3104$\pm$0.00 &0.0814$\pm$0.00\\
FedMVFCM& 0.0177 $\pm$0.00 &0.2645 $\pm$0.00 &0.7896 $\pm$0.00 &0.6479 $\pm$0.00 &0.4912 $\pm$0.00\\
FedMVFPC& 0.0231$\pm$0.00 &0.3623$\pm$0.11 &0.7925$\pm$0.02 &0.8003$\pm$0.03 &0.5166$\pm$0.07\\
FedMSGL& \underline{0.1204}$\pm$0.00 &\underline{0.7543}$\pm$0.03 &\underline{0.9065}$\pm$0.01 &\underline{0.8827}$\pm$0.05 &0.5239$\pm$0.02 \\
\bottomrule
\end{tabular}
\caption{Clustering result on five datasets(mean$\pm$standard deviation)}
\label{result}
\end{table*}

\subsection{Experiment and Parameters Setting}
\subsubsection{Parameters setting} For the compared methods mentioned above, we use the parameters recommended by the authors. For the proposed method FedMSGL, we select $\lambda_1,\lambda_2$ and $\lambda_3$ from the range $\{1e-3,1e-2,\cdots,1e2,1e3\}$. The value of $\beta$ is chosen from the range $\{1e-2,1e-1,1,10,100 \}$. After that, we choose specific hyperparameter combinations based on model performance variations.
\subsubsection{Experiments setting} Centralized multi-view learning methods are directly applied to multi-view datasets. Specifically, the FedMVL and the proposed FedMSGL methods operate within a vertical federated environment. FedMVFCM and FedMVFPC execute within a horizontal federated environment. All experiments are conducted on the i9-14900KF and 32.0GB RAM, MATLAB R2021b. Each method is executed 10 times and the average performance is recorded.

\subsection{Comparison Experiments Results}
We apply three widely used clustering metrics(ACC, Purity and NMI) in the experiments to evaluate the clustering performance of the examined algorithms. The values of these metrics are normalized to 0-1 and a higher value indicates a better performance. Table \ref{result} presents the clustering performance on five datasets. We have underlined the top two algorithms which performed best for clustering. Looking at the big picture,  FedMSGL achieves competitive performance compared with centralised multi-view learning algorithms. Compared with three federated multi-view algorithms(FedMVL, FedMVFCM and FedMVFPC),  FedMSGL achieves significant improvements across five datasets. The results of FSMSC and our FedMSGL excel in most scenarios, indicating that considering both the consistent and view-specific information of multi-view data contributes to better representation. Furthermore, the exceptional performance of FedMSGL across five different types of datasets demonstrates that the learned hypergraph provides a comprehensive representation. Overall, the proposed FedMSGL effectively improves the clustering performances of multi-view learning approach in the federated setting. Detailed analyses of the comparison experiments are presented in the appendix.

\subsection{Sensitive Analysis and Convergence Curve}
Fig.\ref{ps}(a),(b) and (c) illustrate the impact of four hyperparameters on the clustering performance of the FedMSGL method, aiming to investigate the contribution of different components of the model to its performance.  Overall speaking, on the BBC Sport dataset, the local model remains robust to $\lambda_2$, but shows sensitivity to $\lambda_1$ and $\lambda_3$. In the central aggregation phase, the values of $\beta$ should be kept small to prevent the overfitting to the hypergraph imposed structure and ignoring important variations and patterns in the actual data. Fig.\ref{ps}(d) presents the convergence curve of the global model objective function values. We can see that the proposed method decreases rapidly and finally converges to a stable value in a finite number of iterations.

\subsection{Ablation Experiments}
We verify the influence of the hypergraph through the ablation experiments. Due to space constraints, the results of the specific experiments and analyses are included in the appendix. It can be demonstrated that hypergraph provide a considerable improvement in model performance compared to classical graph.

\section{Conclusion}
In this paper, we introduce a novel vertical federated learning method, termed the Self-expressive Hypergraph Federated Multi-view Learning method (FedMSGL). Distinct from existing approaches, our method learns the subspace of uniform dimensions by leveraging self-expressive characteristics, thereby reducing performance loss associated with the inaccessibility of raw data. Furthermore, we employ hypergraphs to capture complex cross-view relationships, culminating in a comprehensive global consistent model. Empirical evaluations on real-world datasets demonstrate that our proposed algorithm achieves competitive performance. These results highlight the potential of FedMSGL to effectively address the unique challenges posed by multi-view data in federated environments.

\newpage
\section{Acknowledgments}
This work was supported in part by the Guangdong Basic and Applied Basic Research Foundation under Grants 2022A1515010688 and in part by the National Natural Science Foundation of China under Grant 62320106008. 

\bibliography{mylib}

\end{document}